# ISOMETRIC LINEATION IN ENGLISH TEXTS

## AN EMPIRICAL AND MATHEMATICAL EXAMINATION OF ITS CHARACTER AND CONSEQUENCES

Hideaki Aoyama & John Constable

FACULTY OF INTEGRATED HUMAN STUDIES
KYOTO UNIVERSITY

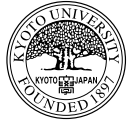

1998

*Abstract*

In this paper we build on earlier observations and theory regarding word length frequency and sequential distribution to develop a mathematical characterization of some of the language features distinguishing isometrically lineated text from unlineated text, in other words the features distinguishing isometrical verse from prose. It is shown that the frequency of $Q_n$ of $n$ syllables making complete words produces a flat distribution for prose, while that for verse exhibits peaks at the line length position and subsequent multiples of that position. Data from several verse authors is presented, including a detailed mathematical analysis of the dynamics underlying $Q_n$ peak creation, and comments are offered on the processes by which authors construct lines. We note that the word-length sequence of prose is random, whereas lineation necessitates non-random word-length sequencing, and that this has the probable consequence of introducing a degree of randomness into the otherwise highly ordered grammatical sequence. In addition we observe that this effect can be ameliorated by a reduction in the mean word length of the text (confirming earlier observations that verse tends to use shorter words) and the use of lines varying from the core isometrical set. The frequency of variant lines is shown to be coincident with the frequency of polysyllables, suggesting that the use of variant lines is motivated by polysyllabic word placement. The restrictive effects of different line lengths, the relationship between metrical restriction and poetic effect, and the general character of metrical rules are also discussed.

# 1. Introduction: Verse and Prose

When Auden wrote that "The difference between verse and prose is self-evident" (1963: 23), he was not suggesting anything controversial, and would presumably have agreed with Christopher Ricks that the variation presented "no interesting puzzle" (1970: 261). Yet, as Northrop Frye has observed, this "most far-reaching of literary facts" is a "distinction which anyone can make in practice" but which "cannot be made as yet [...] in theory" (Frye 1957: 13). There has been much discussion since Frye's remarks, and some progress, T. V. F. Brogan even citing this passage in his *Princeton Encyclopedia of Poetry and Poetics* entry, "Verse and Prose", to illustrate the distance now separating the poetics of the mid-Fifties from that of the early nineties (1993c: 1346). However, as the rest of that article demonstrates, no clear theoretical distinction has in fact been articulated in detail, and "Distinguishing between prose and verse remains as perplexing an issue as ever" (Steele 1990: 81: see also Marks 1998: 148). This is somewhat surprising, since the elements for a satisfactory theory have long been available, at least since T. S. Omond remarked that difference was one of "mechanical method":

> The units of prose are diverse, irregular in length, rarely conformed to a common pattern. In verse, on the other hand, succession is continuous. Something recurs with regularity. (Omond [1903], quoted in Matthews 1911: 31)

Lotz (1972) puts this general point in a still more satisfactory way:

> Verse and prose are opposed to each other as two types of which, one, verse, has definable properties, and the other, prose, is characterized by lack of any such features. (1972: 1)

> In most languages there are texts in which the phonetic material within certain syntactic frames, such as sentence, phrase, and word, is numerically regulated. Such a text is called verse, and its distinctive characteristic is meter. [...] A non-metric text is called prose. Numerical regulation may refer to a variety of phenomena; therefore, verse and prose are distinguished not as two sharply differentiated classes, but rather as two types of texts. (This, however, should not obscure the fact that verse and prose are polar opposites [...].) (1972: 4-5)

> [...] the deviation [of verse from prose] can be put in terms of numerical regularity, or meter, and this regularity is the *differentia specifica* of verse (1972: 6)

Nevertheless, to our knowledge, there is neither empirical work demonstrating the existence of these regularities, nor theoretical work examining their mathematical character.

We will attempt to remedy this situation, and clear the way for a more productive analysis, by offering theory and data bearing on the distinction between prose and isometrically lineated text in English. Lineation, as may be inferred from Lotz's principles, is a basic feature of meter, and there is now general agreement that the arrangement of text in lines is the most commonly occurring rule in the world's metrical systems (Brown 1991: 132; Lotz 1972: 19; Tarlinskaia 1989: 122). Indeed some scholars have gone further and suggested that there are good reasons for accepting the cultural "universality of the line" (Dell Hymes, quoted in Leavitt: 1997: 134). However, even exceptionally thorough literary linguistic manuals content themselves with only brief discussions of verse lineation (Leech

1969: 114; Williams 1986: 182-187; Tarlinskaia 1993: 3-5; Gasparov 1996: 1-3; Fabb 1997: 88-91), and attention has been mostly directed onto the admittedly very interesting question of rhythm within lines (Attridge 1982; Attridge 1995). In the narrower field of English literary criticism, the phenomenon has drawn much comment relating to its effects on intepretation (see Culler 1975: 183ff for a discussion of several interpretative strategies by which line breaks are "accorded some kind of value"), but little detailed examination of its nature. When not engaged with subtle interpretational questions, critics locate their analysis at a high level of abstraction, and range over and attempt to conjoin the concepts of isometric lineation, the lineation of free verse, and even poetic effect itself, a very broad range of subjects (Holder 1995: 137ff provides a useful review of several prominent authors). This is an extremely ambitious project which has served to show the interest of the issues involved, but has not, in our view, been successful in deepening scholarly understanding of the questions under consideration.

The last of these, poetic effect, has been the subject of such a long-standing and extensive literary critical debate on the differentia of *poetry* and prose (for examples see Darwin 1789: 40-41; Whately [1828] 1963: 333-334; Newton Scott 1904; Lotspeich 1922; Murry 1922: 47-70; Read [1928] 1949: ix-xiii; Alexander 1933: 84ff; Furniss and Bath 1996: 12-13; Leonard 1996: 75) that we should, perhaps, make it clear at the outset that we accept the substance of Wordsworth's remarks that "much confusion has been introduced [...] by this contradistinction", and believe him to be correct when he writes that the "only strict antithesis to prose is verse" ([1800] 1974: 134. See also Hamer 1933: 1, and Turco 1986: 5 for helpful remarks on this issue).[1] This is not to slight the importance of the relationships between prose, verse, and poetic effect, but we submit that consideration of the third term and its relations must wait on better understanding of the differences between the first two. Consequently, we will concern ourselves initially with the question of illuminating those features which make regularly lineated text distinct from unlineated, and only then will we turn, briefly, to the vaguer and more problematic question of rich poetic effects and their peculiar association with both prose and verse.

---

[1] Wordsworth subsequently revised his text, arguing that since "lines and passages of metre so naturally occur in writing prose" the division of prose and verse was not "in truth, a *strict* antithesis" ([1850] 1974: 135; see also Brooks and Warren 1960: 122, who come to similar conclusions). As pointed out by Lotz (1973: 5) the fact that prose and verse are not distinct classes should not obscure the fact that they are distinguishable types.

By isometric lineation we understand a requirement that lines be of the same length, and we shall focus on length measured in syllables, though this is in fact only indirectly specified through regulation of the number of beats and offbeats. Variations in length result from flexibility in the realization of the beat/offbeat pattern, and an English text composed in ten syllable lines, for example, usually displays a number of nine, eleven and twelve syllable lines in addition to its core isometrical set. Further, we include within the isometric category repetitions of heterometrical structures. For example the Spenserian stanza, which is a structure of eight decasyllabic lines and a concluding twelve syllable line. We will not discuss heterometrical lineation, that is to say non-stanzaic verse where there are lines of various lengths and no repetition (Brogan 1993a), for the simple reason that our methods are not sufficiently sensitive to detect lineation in small samples. Of the lineation of free verse, we will say almost nothing, since, as will be made clear later, it lies beyond the scope of our investigation.

Most readers and writers of English verse are aware that isometrically lineated text is not merely a visually displayed distribution of words (Scott 1979: 158) but is in some sense a matter of the language. Here and there in the literature there are numerous remarks which touch on this matter. Levin, for example, notes that "Except possibly in free verse, the typographical groupings [...] are not random; some organizing principle must thus be at work behind them" (1960: 181).[2] Chatman has observed that "The poet may select his words in part by considering the numbers of syllables they contain" and thus that "number becomes a part of the poem's mode of existence" (1970: 318-319). Lotz advances the discussion by noting the importance of the regular occurrence of word boundaries (1972: 7-8), a point also made by Jakobson (1960: 361) in relation to Russian verse, and repeated by Fabb (1997:88). Rothman, in a defense of the metrical status of pure syllabic verse, cites the word boundary rule, and very pertinently observes that lineation thus exercises some degree of control over the "phonemic flow" in the text (1996: 207). All these remarks have something to recommend them, but it is in Wimsatt and Beardsley's paper "The Concept of Meter" that we find the most suggestive sketch of the heart of the matter:

> [...] to have verses or lines, you have to have certain broader structural features, notably the endings. Milton's line is not only a visual or typographical fact on the page, but a fact of the language. If you try to cut up his pentameters into tetrameters, for example, you find yourself ending in the middle of words or on weak words like 'on' or 'the'. Much English prose is iambic or nearly iambic, but it is only very irregular verse, because if you try to cut it regularly, you get the same awkward and weak result. (1959: 591)

---

[2] Some readers may be surprised to see Levin's article cited in support of this view, since he elsewhere in the same piece states that "the line is a purely typographical device" (179), and in the colloquium discussion which followed the conference presentation of the paper (193-196) was taken to task for suggesting that lineation was not "not linguistic". Careful reading of Levin's text, however, reveals that his definition of "linguistic" is a scrupulously restricted technical use, and that in his terms he can claim that lineation is non-linguistic, that it is conventional, while not calling into question the fact that lines "comprise patterns or structures of language elements".

While we do not wish to defend the value of raising the "weak words" issue in a discussion of the fundamentals of lineation (it is a supplementary matter in our view), or the doubtful claim that English prose is iambic, the proposition that the relation of polysyllables to line boundaries is crucial in understanding the character of lineated text deserves more attention than it has hitherto received, and we will attempt to develop this insight, and those of the other authors cited above. Initially, our examination concerns itself with the phenomenon as a feature of completed texts, and then in subsequent sections we will change perspective and look at the question from the author's viewpoint, asking what sort of syntactical and dictional choices are forced or encouraged by lineation. It may be observed in passing that though the thesis presented here is exclusively concerned with the phenomenon of lineation in English we anticipate that many of our remarks may be abstracted and applied to other languages, though it seems probable that adjustments will be necessary to take into account features peculiar to those languages.

## *2. Mathematical Characteristics of Unlineated and Lineated Text*

As noted above, an understanding of the characer of lineation involves an examination of the distribution of polysyllables in output, and this entails a discussion of both frequency totals and sequential distribution. In this section we will summarize work of our own on prose and demonstrate that verse is significantly different in its sequential distribution.

## 2.1 Unlineated Text: Geometric Distribution, Random Sequencing and the Flat $Q_n$ Distribution

In a previous article (Aoyama and Constable In Preparation) we have analyzed the word-length structure of almost two million words of prose. By using a symbol manipulation program we computed the frequency of *all* sequences of complete words totalling one syllable, all sequences of complete words totalling two syllables, and so on throughout the text up to complete words totalling thirty syllables. Thus, we determined the number of matches to the line definition rule given by Constable, stating that a line of a specified length "*must only be complete words*" (1997: 181). The normalized frequency $Q_n$ of such occurrences is defined by the equation:

$$Q_n \equiv \frac{L_n}{I}, \qquad (1)$$

where $L_n$ is the number of matches to a line definition rule of $n$ syllables, and $I$ is the total number of syllables in the data, and was found to be in general flat (the boundary condition was resolved by connecting the end of the data with the beginning), that is to say that the $Q_n$ is independent of $n$ (see Fig. 1; the same data is given in an enhanced form in Fig. 2), in agreement with remarks made earlier by Constable (1997: 182).

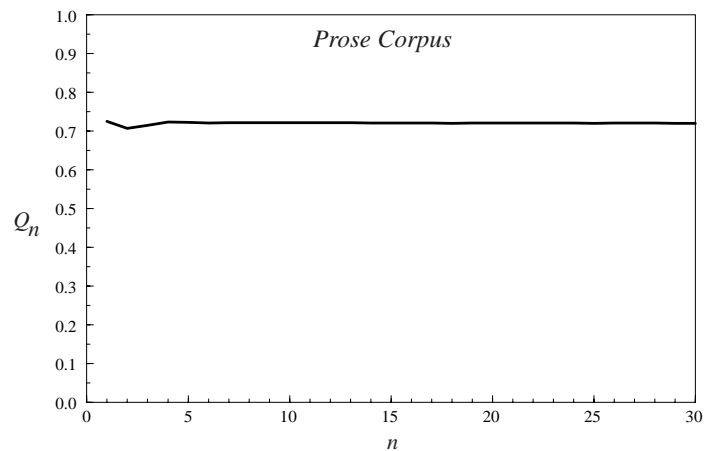

Figure 1: Plot of $Q_n$ for all the data in the prose corpus, containing 1,977,676 words. It is evident that its single unique characteristic is its flatness, that is the independence of $Q_n$ from $n$.

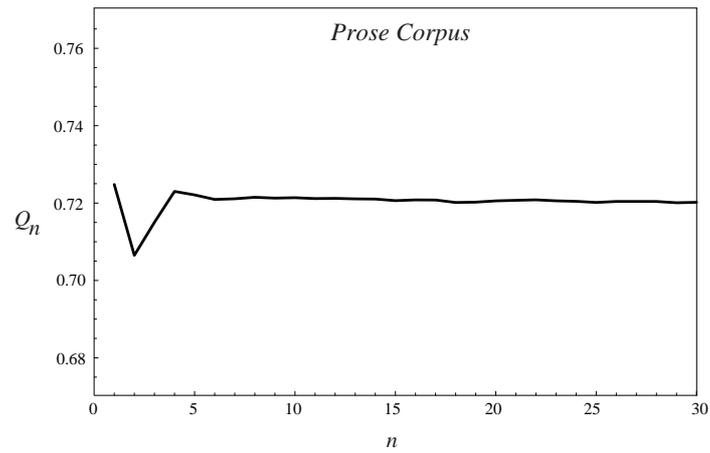

Figure 2: Detail of Fig. 1. Note that the vertical scale for $Q_n$ is enhanced ten times.

This flat distribution is also characteristic of each text used in the corpus, and does not result from averaging over individual variations. Figs. 3 and 4 represent data from George Eliot's *Middlemarch* and Henry James' *The Portrait of a Lady*, respectively.

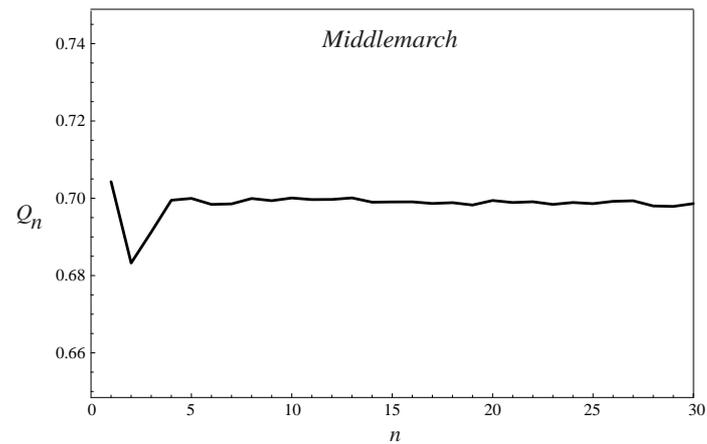

Figure 3: The $Q_n$ distribution of *Middlemarch*, containing 317,827 words. This is an enhanced view, as in Fig. 2.

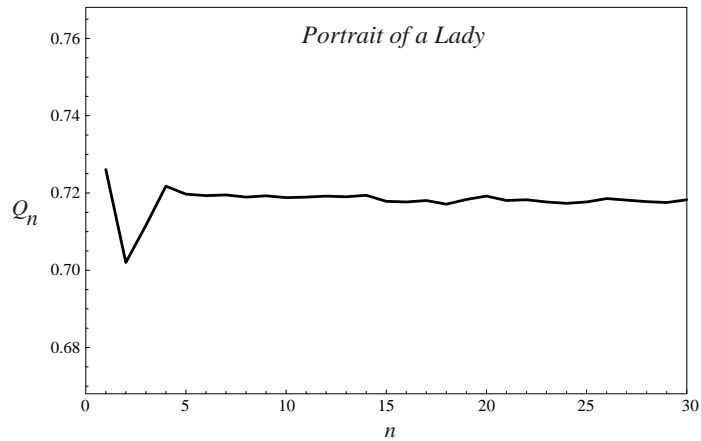

Figure 4: The $Q_n$ distribution of *Portrait of a Lady*, containing 225,354 words, in the enhanced view.

It was also observed that there is in general no correlation between the syllable counts of adjacent words, that is to say that the number of syllables in a word has no effect on the value of the subsequent word (see Fig. 5). When taken together with the flatness of the $Q_n$ distribution this random ordering implies a geometric distribution of word length frequency totals, which can be empirically confirmed from the data (see Fig. 6). This leads us to the conclusion that *prose is randomly segmented*.

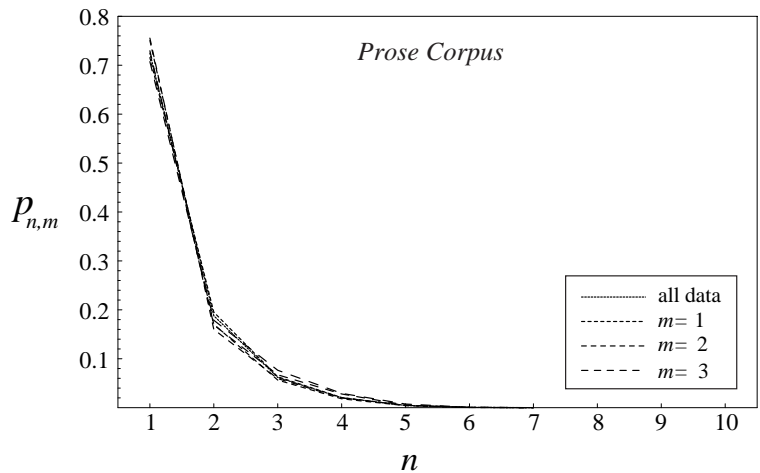

Figure 5: The probability distribution $P_{n,m}$ of having an $n$-syllable word after an $m$-syllable word for the prose corpus.

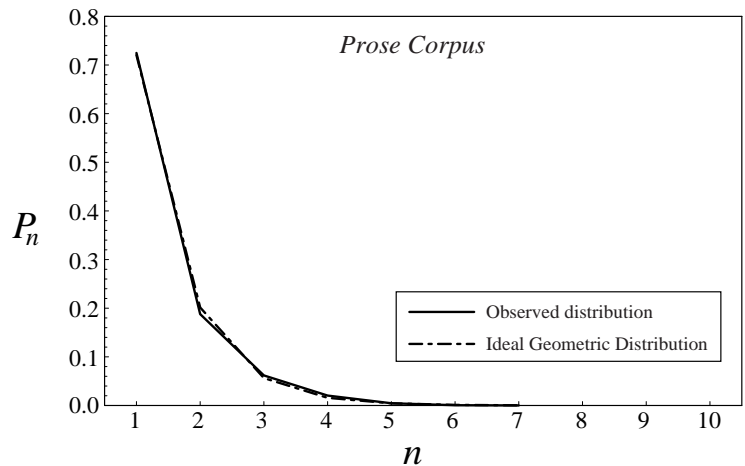

Figure 6: Observed probability distribution $P_n$ of the number of syllables in a word being $n$ for the prose corpus, compared with an ideal distribution.

Readers will have noticed that the beginning of the $Q_n$ distribution shows minor deviations from the features mentioned above, particularly the depression at $n = 2$, and in fact some of these are of statistical significance. However, they are not relevant to the present stage of our analysis, though we will refer to their importance in a subsequent section.

## 2.2 Lineated Text: $Q_n$ Peaks in Isometric Verse, Empirical Studies

We have now demonstrated that the sequence of word length items in unmetred output, prose, is a random sequence of items from a geometric frequency distribution, and that the $Q_n$ frequencies produce a flat distribution. Lineated text differs considerably from unlineated text in this respect. We have found distinct peaks at the position where $n$ is equal to the most probable number of syllables per line, and at subsequent multiples of that line length (in texts where there are a number of variant line lengths, as there often are, these peaks progressively diminish in prominence).

We have analysed seven verse texts (details are given in Table 1.), the syllabic data being obtained with a simple marking program, written by one of us (Constable), which uses a custom built lexicon to determine the syllabic count of each word. The texts were either obtained from public on-line sources or prepared by us. $Q_n$ charts are given in Figs. 7 to 13.

| Author | Text | Date | No. of Words | Line Length Distribution | | | | | | | |
|---|---|---|---|---|---|---|---|---|---|---|---|
| | | | | 7s | 8s | 9s | 10s | 11s | 12s | 13s | 14s |
| Edward Fairfax | Jerusalem Liberata | 1600 | 124,856 | 3 | 45 | 1,027 | 12,653 | 1,632 | 112 | 7 | 0 |
| George Chapman | The Odyssey | 1616 | 68,170 | 0 | 3 | 227 | 7,217 | 1,073 | 67 | 1 | 0 |
| John Milton | Paradise Lost | 1667 | 79,836 | 0 | 1 | 163 | 8,315 | 1,887 | 178 | 4 | 0 |
| John Dryden | The Aeneid | 1693 | 106,483 | 0 | 1 | 86 | 11,544 | 1,483 | 515 | 65 | 5 |
| Mark Akenside | Pleasures of Imagination | 1744 | 14,512 | 0 | 0 | 0 | 2,006 | 1 | 0 | 0 | 0 |
| William Wordsworth | The Prelude | 1839 (D text) | 57,570 | 0 | 1 | 46 | 6,087 | 1,523 | 181 | 10 | 1 |
| H. W. Longfellow | The Song of Hiawatha | 1855 | 31,038 | 68 | 5,021 | 314 | 9 | 1 | 0 | 0 | 0 |

Table 1. Verse Texts Analysed: Line Length Distributions

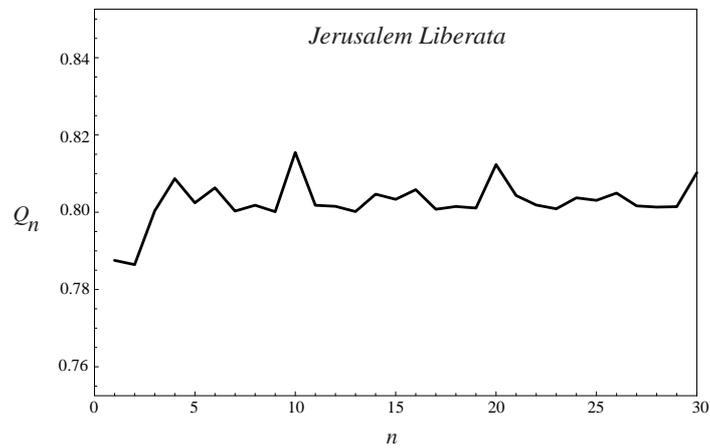

Figure 7. Edward Fairfax, *Jerusalem Liberata*: Detailed $Q_n$ Distribution

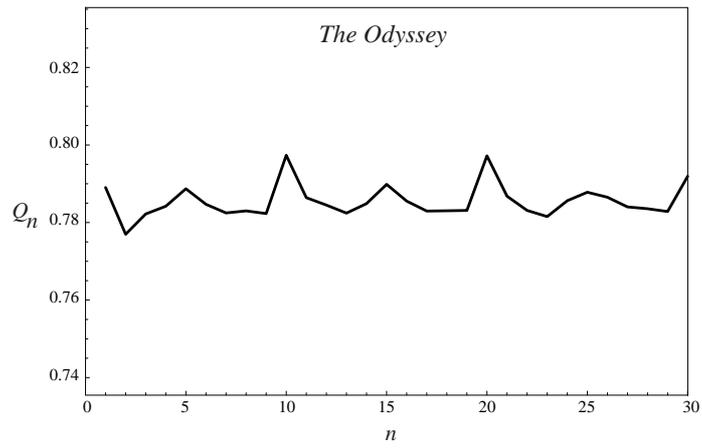

Figure 8: George Chapman, *The Odyssey*: Detailed $Q_n$ Distribution

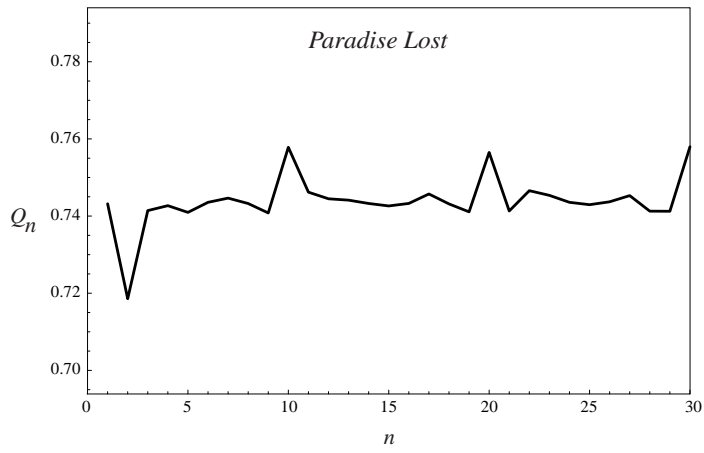

Figure 9: John Milton, *Paradise Lost*: Detailed $Q_n$ Distribution

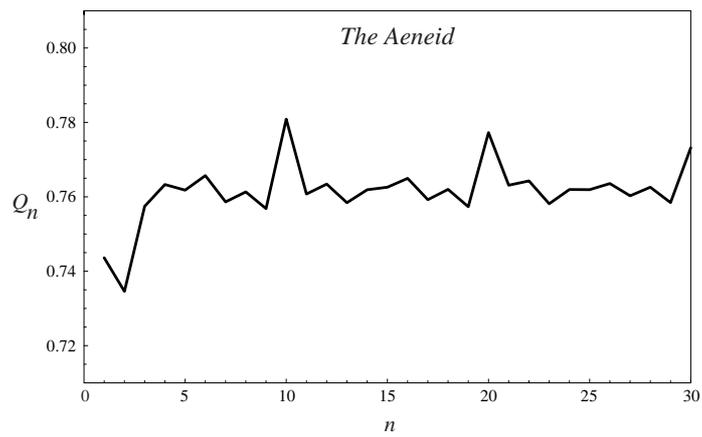

Figure 10: John Dryden, *The Aeneid*: Detailed $Q_n$ Distribution

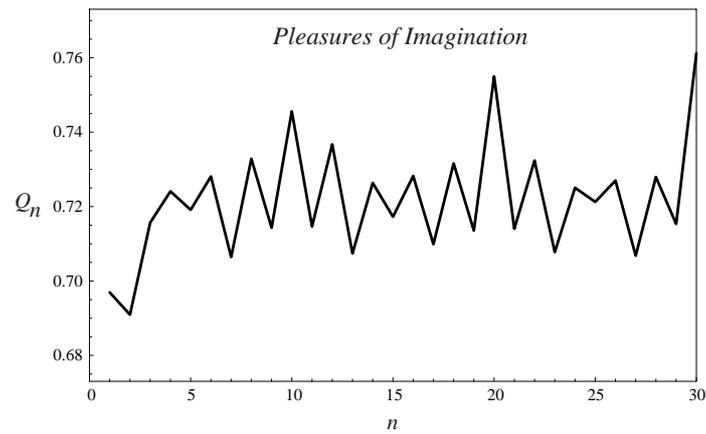

Figure 11: Mark Akenside, *Pleasures of Imagination*: Detailed $Q_n$ Distribution

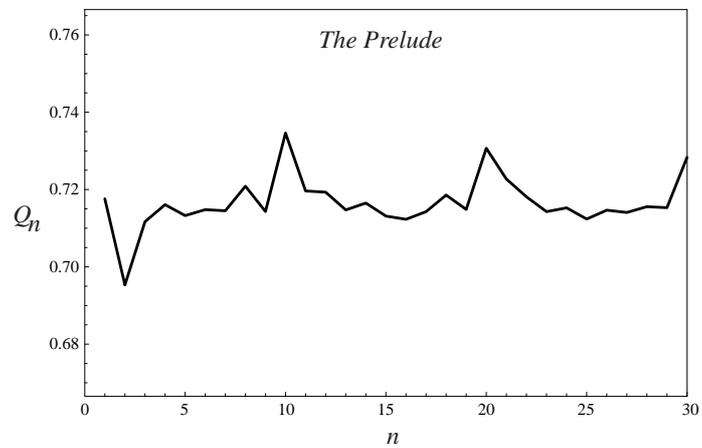

Figure 12: William Wordsworth, *The Prelude*: Detailed $Q_n$ Distribution

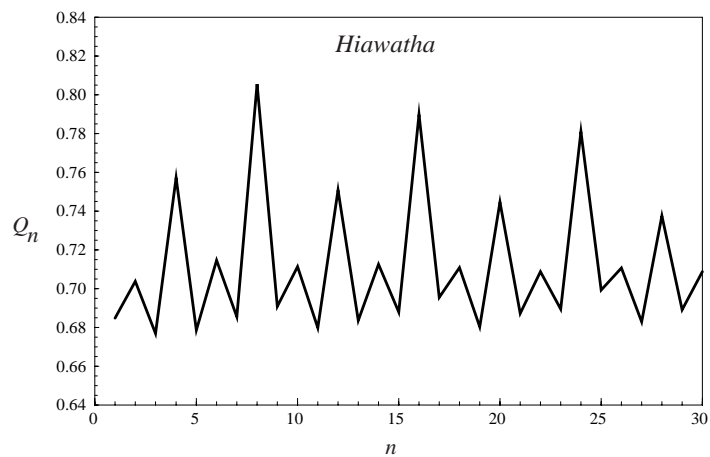

Figure 13: H. W. Longfellow, *The Song of Hiawatha*: Detailed $Q_n$ Distribution. Note that for this particular plot, the vertical scale covers the range of width 0.2 in $Q_n$, twice that employed in the other detailed figures.

The results for, Fairfax, Milton, Dryden, and Wordsworth are straightforward. There are significant peaks over each of the most probable line lengths. It will be also be noticed that the figures for Longfellow, and, for Chapman, are more complex, since they give obvious evidence of a regularity of segmentation at lengths lower than that graphically represented on the printed page. That is, in the case of Longfellow, four syllable groups are more common than would be expected by chance, and in the case of Chapman the same is true of five syllable groups. Longfellow is the most singular case, having regular and substantial oscillations. These structures of the $Q_n$ distribution can be understood from the underlying structures of the segmentation within each line as we will discuss in the following section. In fact, we will show that the examples of Chapman and Longfellow, where the patterning is readily observed in the charts, are extreme cases, and that the internal structure of the lines, that is the subsidiary peaks between the main $Q_n$ peaks in all the texts, including the Akenside, where it might be thought that the noisiness of the chart results from the small sample size, is to varying degrees non-random, the cause of this structuring being found in the effects of beat patterning, rather than in lineation itself. It seems probable that this will be a profitable area for future research, particularly when considering differences between authors.

Overall, the importance of the $Q_n$ peaks should be immediately apparent; in addition to the visual and aural line (Brogan 1993b: 696) we can now speak of the mathematical line. In other words, the view of authors who have supposed that lineation is arbitrary in terms of language, a rare position in its pure form (but see Weirather 1980, quoted in Holder 1995: 136) though prevalent as a general uncertainty as to the ontological status of lines, is shown to be untenable. The claims of Wimsatt and Beardsley, and the intuitions of the other authors cited above, are correct. Lineation is a fact of the language.

The concrete demonstration of this point may bring some order to the debate around the issue of the relation between sound and lineation (see Holder 1995: 129-155). Some critics have wondered, for example, whether Milton's blank verse was "verse only to the eye" (Johnson 1967: 193; Bradford 1988 summarizes the positions of several other seventeenth and eighteenth century authors), and others have argued that pure syllabic verse is not aurally significant, and therefore "not a metre in English" (Baker 1996: 12). That lines are often perceived aurally, a position taken by many authors, for example by Hopkins ([1874] 1959), and endorsed very prominently by Jakobson (1960: 358), should not be in question. However, our findings show that the salience of lineation for readers may not be of primary relevance to understanding the character of lineated text. Consequently, Johnson's skepticism, and Wallace's claim that syllabic verse is a "kind of free verse" (1996: 12) seem to us to be confusing two distinct issues, on the one hand the salience of lineation, and on the other its ontological status. It seems to us that most scholars have thought they were addressing the latter question, when in fact they were discussing the ease with which readers or listeners might recognize lineation if they were not able to use the typographical appearance of the page to assist them (Gioia 1996 and Rothman 1996 make similar points in their reply to Wallace 1996). Sonic rhythm, and rhyme, undoubtedly work to mark line endings, as do carefully positioned syntactic boundaries. However, even if a text were composed in isometric syllabic lines without sonic markers, which might be very difficult indeed to recognize if printed in an undisplayed form, there would still be a real mathematical distinction between the language of such a text and that of a prose text. Whether this suggests that there is an "essential difference between the language of prose and metrical composition" (Wordsworth [1800] 1850) is largely a matter of terminological definition; if word length sequencing is deemed an "essential" characteristic of language, then there is an essential difference between prose and verse, if not, then there is none. We will content ourselves with pointing out that there is a computable difference, and leave it to the reader to decide whether this is an essence or not for any particular purpose.

It might be suggested that our procedure constitutes a "test" for lineation, assisting in, for example, analysis of authors known to have printed verse as prose, such as Melville, Scott Fitzgerald, and Dickens. However, we would caution against this, since our technique is not sufficiently sensitive to detect the presence of short lineated passages within a large body of prose, where any $Q_n$ peak would be so small as to be effectively concealed by random fluctuations. However, the general principle that we have outlined might function as a test in cases where there is doubt as to the metrical status of relatively lengthy texts or substantial canons. For example we suggest that with suitable adjustments our approach might contribute something to the discussion of Hebrew biblical texts, the line construction principles of which have been much debated (Yoder 1972; Kugel 1990; Geller 1993). Of course, the language features of Hebrew may be so different that our method is not applicable, and even if it is appropriate it may be that difficulties in establishing reliable texts and syllable counts would stand in the way of productive research (Yoder 1972: 59).

## 3. Composing in Lines and $Q_n$ Peak Construction

We have now shown that isometrically lineated text is marked by significant empirical properties differentiating it from unlineated text. While the mathematical dynamics underlying these peaks could be approached in an abstract manner, the analysis can be more firmly motivated by placing it within a consideration of the compositional process. It should be noted at the outset that the discussion below focuses on duple verse, that is verse where the offbeat position is typically occupied by a single syllable, and that triple verse, where the offbeat is typically or very much more frequently a pair of syllables, requires a separate and presumably very different discussion.

### 3.1 Theoretical Line Violation and Alternative Placement

If a passage of prose is arbitrarily broken into lines at every tenth syllable a certain number of these breaks will lie within polysyllabic words. Since the sequential distribution of word length items is random, if we know the word length frequency totals we can predict how many of these introduced breaks will actually lie within words. For example, George Eliot's *Middlemarch* contains 317,827 words, and 456,620 syllables, with a word length frequency distribution described in Table 2. The number of syllable boundaries internal to words is calculated in column four, and column five presents this information as a proportion of all syllable boundaries in the text.

| Word Length | Frequency | Normalized Frequency | Number of Syllable Boundaries internal to words | s/bs internal to words as proportion of all syllable boundaries |
|---|---|---|---|---|
| 1 | 223,840 | 0.704 | 0 | 0 |
| 2 | 61,588 | 0.194 | 61,588 | 0.135 |
| 3 | 22,256 | 0.070 | 44,512 | 0.097 |
| 4 | 8,118 | 0.026 | 24,354 | 0.053 |
| 5 | 1,806 | 0.006 | 7,224 | 0.016 |
| 6 | 201 | 0.001 | 1,005 | 0.002 |
| 7 | 17 | 0.000 | 102 | 0.0002 |
| Total | 317,826 | 1.000 | 138,785 | 0.3032 |

Table 2: George Eliot, *Middlemarch*: Proportions of Syllable Boundaries Internal to Words

This calculation can be done for the ideal geometric distribution as follows. In a previous paper (Aoyama and Constable, In preparation), we have defined the following constant frequency distribution:

$$\overline{Q}_n = q. \tag{2}$$

It was shown that this is induced by the geometric distribution for the syllable counts of each word, $\overline{p}_n = q(1-q)^{n-1}$. The probability of a syllable boundary being internal to a word is given by

$$\sum_{n=1}^{\infty}(n-1)\overline{p}_n \bigg/ \sum_{n=1}^{\infty} n\overline{p}_n = (1-q) \tag{3}$$

This result leads us to an obvious interpretation: By the definition (2), $q$ is the probability of a given syllable bounday being a word boundary. Since the above probability is the case opposite to this, it is given by (2). In fact, the *Middlemarch* data has an average $Q_n$ of about $\langle Q \rangle = 0.6961$, and using this as the estimate of $q$, we find that this theoretical estimate is $1-q = 0.3039$, which is in good agreement with the value 0.3032 obtained in Table 2.

Returning to our concrete example, let us suppose that the text is broken into 45,662 ten syllable lines. We know that 0.3 of all syllable boundaries are internal to words, and thus, because words are sequentially distributed at random, we also know that, probabilistically, 0.3 of the line boundaries will fall within words. Thus we can see that approximately 13,700 words will violate line boundaries. That is to say, if this text were to be versified, or if an author were to attempt to construct verse in similar language, then the composer would be faced with the task of moving 13,700 words away from the line boundary positions. This is the core of the act of lineation, and constitutes the informational ordering entailed by that act, and is registered in our procedure as the $Q_n$ peak. It should be noted that longer lines result in fewer theoretical violations, and therefore result in less work for the author, as do texts with lower proportions of polysyllables. It should also be noted at this point that the size of the $Q_n$ peak is, as might be expected, related to the number of theoretical violations, and thus to both mean word length and line length (a fuller discussion of this point will follow in the subsequent discussion of Eq. 14).

We have now seen that lineation is fundamentally a process whereby polysyllables are placed non-randomly with regard to line boundaries. That is to say as a text is composed an author will from time to time, for communicative and grammatical reasons, be inclined to place a polysyllable in a line boundary violating position, and we can think of lineation as the act of avoiding such violations. There are two techniques, both of which create $Q_n$ peaks: 1.

alternative placement, and 2. word length reduction. In addition, when these techniques fail to produce satisfactory results authors will vary from the core metrical line length, producing variant lines, thus reducing the $Q_n$ peak size. We will discuss each of these practices in turn.

### 3.1.1 Alternative Placement: Non-Random Sequencing Within Lines

Rothman has suggested that in pure syllabic metre "there is a pleasing and challenging strain between numbers of syllables and numbers of words to a line", a point which we endorse, and, further, that "at the end of each line word choice becomes highly constrained by syllable count" (1996: 207). We take this as suggesting that authors tend to make line-fitting adjustments to their word length towards the end of the line. This point is applicable by extension to other line forms, such as those we are discussing, where line length regulation is the outcome of beat and offbeat regulation. We can evaluate this hypothesis by examining the distribution of words of varying numbers of syllables within each line, a question which can best be approached by determining the frequency of spaces following any position in the line. If there is any tendency for words in a region to be longer, then the frequency of spaces in that region will also decline. Figs. 14 through to 20 chart data for the seven texts listed in Table 1. Figures relating to the core metrical line (dots and solid line) and the most frequent variant (squares and dotted line) are reported. The horizontal axis $n$ represents syllable positions and the vertical axis $x_n$ represents the normalized frequency of spaces following these positions. There appears to be no tendency for words at the end of the line to be shorter, indeed if anything there is a tendency for them to be longer, though the overall distribution is globally even.

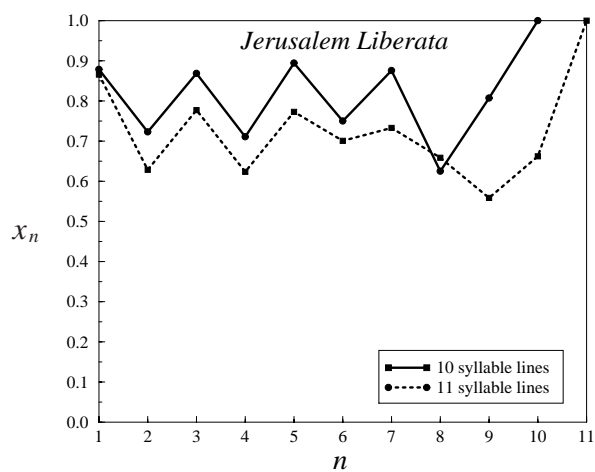

Figure 14: Edward Fairfax, *Jerusalem Liberata*: Distribution of Spaces Within Lines

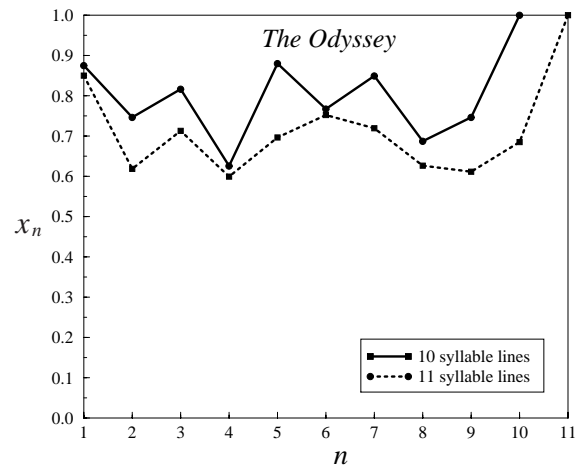

Figure 15: George Chapman, *The Odyssey*: Distribution of Spaces Within Lines

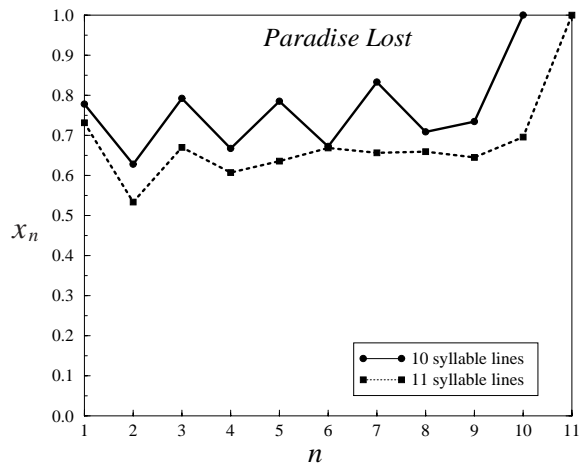

Figure 16: John Milton, *Paradise Lost*: Distribution of Spaces Within Lines

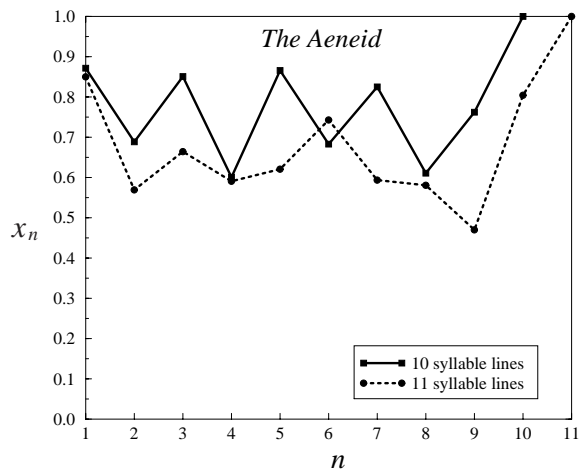

Figure17: John Dryden, *The Aeneid*: Distribution of Spaces Within Lines

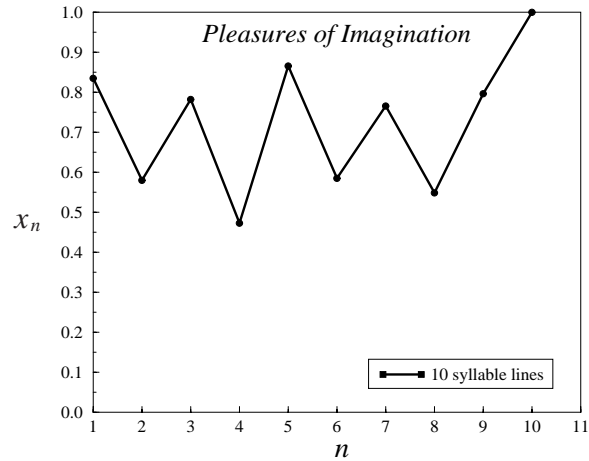

Figure 18: Mark Akenside, *Pleasures of Imagination*: Distribution of Spaces Within Lines

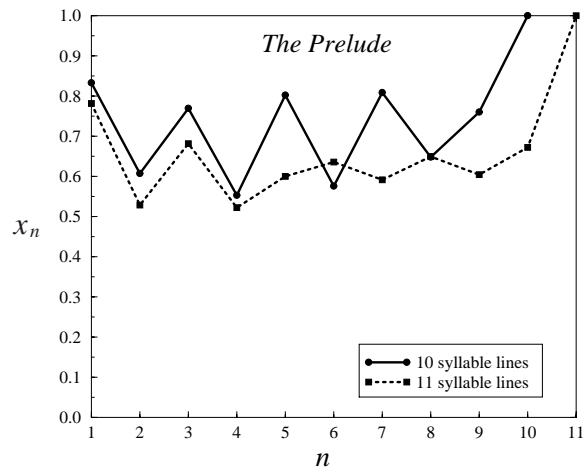

Figure 19: William Wordsworth, *The Prelude*: Distribution of Spaces Within Lines

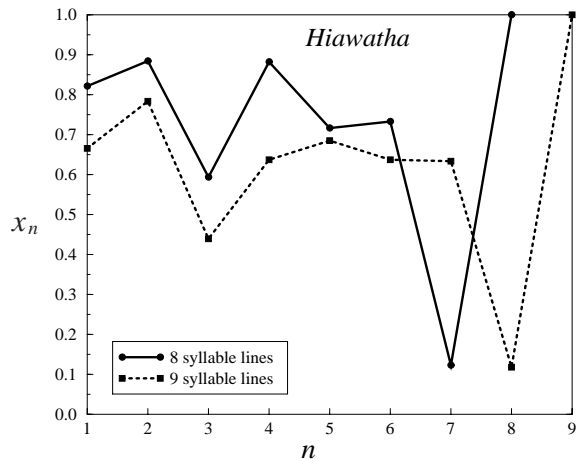

Figure 20: H. W. Longfellow, *The Song of Hiawatha*: Distribution of Spaces Within Lines

## 3.1.2 Word Length Sequencing Within Lines: Concluding Observations and Additional Remarks

We have already noted that the overall distribution of word length is relatively even, however, our texts are all by reputable authors, presumably highly skilled in the craft of verse, and it seems possible that less able writers may not achieve such an even distribution of word length throughout the line. Further, Constable (Forthcoming a) has argued at length that revisions to verse will tend to give evidence of increasingly successful solutions to the problems posed by metrical form, and it would be of great interest to know if revised verse texts tend towards a smoother distribution. Constable (Forthcoming a) has already shown that Akenside's revision of *Pleasures of Imagination* uses a much greater variety of structures within lines, and we predict that this will be found to be an indirect indication of this smoother word length distribution.

The space distribution charts above (Figs. 14-20) also give evidence for two further observations:

- For the most frequent line lengths, the distribution follows an alternating pattern, with enhancements for the odd numbered syllables in all cases except *Hiawatha*, where the enhancement occurs for the even-numbered syllables.
- For the next-to-most-frequent lines, the distribution is stretched toward the end of the line. That is to say that it tends to be flat toward the end of the line, especially after the medial syllable.

The first of these points clearly suggests some degree of non-random ordering of word length, and the second, though peripheral to the main topic under discussion here, is so prominent that it requires explanation.

*Alternating Pattern*

Two facts may be employed to explain the phenomenon of the alternating pattern whereby we see that spaces are much more common after certain positions in the lines (this observation is, of course only valid for the duple rhythm verse we have examined; triple rhythm may display very different effects). Disyllables, which make up the bulk of polysyllables in any text, are usually stress initial. In a 17,000 word dictionary compiled by one of us (Constable) from a range of texts in prose and verse 0.77 of all disyllables, of which there were approximately 7,500, were found to be of this form. Secondly, in English verse lines stresses in polysyllables are almost invariably required to coincide with one of the beat positions. Consequently, there are more spaces before beat positions than after them. In most of the texts considered the beats occur on the even syllables, except in *Hiawatha* where they occur on the odd syllables, and thus we find a strong alternating pattern of space frequency.

*Decay of the Alternating Pattern in Variant Lines*

A ten syllable five beat line is made up of five beats and five offbeats, and can be described thus, where O represents offbeats and B represents beats:

OBOBOBOBOB

Each beat is a single syllable, each offbeat is a single syllable. The O positions are 1,3,5,7,9, the B positions are 2,4,6,8,10. (There are some possible variations from this pattern, BOOBOBOBOB for instance, but these, though interesting, are not common enough to detain us here.)

Variant lines which are longer than this pattern will employ one or more double offbeats, that is offbeats containing two syllables. In the eleven syllable line this will be a single double offbeat, placed in one of the five possible offbeat positions. When this occurs the beat and offbeat positions are all shifted rightwards by one value. That is, if the double offbeat occurs in the first O position, then the beat positions subsequently become 3,5,7,9,11, and the offbeat positions 4,6,8,10. Obviously, only the positions subsequent to the double-offbeat are shifted, and the preceding positions, if any, are unchanged.

Now, for the purposes of discussion we may assume that the double offbeats in the eleven syllable lines are evenly distributed over the five positions (in fact some authors may prefer certain locations to others, but we will ignore this for the time being). Thus we should expect in our sample that 0.2 of all the lines have a double offbeat in the first offbeat position, 0.2 in the second and so on. Thus, we should expect that 0.2 of the lines are right-shifted subsequent to the first offbeat position, 0.4 of the lines are shifted subsequent to the second offbeat position, 0.6 after the third, 0.8 after the fourth, and all them will be shifted by the line end.

The implications for the alternating pattern are clear. It will be more or less intact for the first few positions, but will decay as the line progresses. In the first positions most of the lines still have beats on the even numbered positions, but later in the line more of them will be shifted. Indeed, it is reasonable to assume that a shifted alternating pattern will begin to strengthen towards the end of the line, and this appears to be detectable in the charts. With this digression behind us we will now return to the question of lineation.

At this point it becomes reasonable to ask what relationship exists between the $x_n$ and the $Q_n$ distributions. In the following sections we will show that the non-random $x_n$ distributions within lines in fact account for the subsidiary peaks seen in the $Q_n$ distributions.

### 3.1.3 $Q_n$ Distribution induced by the $x_n$ distribution

We will first investigate the relation between the space distribution in lines, and the $Q_n$ frequency distribution. Let us assume that the text is made of only one type of the core metrical lines of length $N$. If the lines of various pattern types are distributed randomly in this text, the $Q_n$ distribution is expressed as follows, using the $x_n$ distribution,

$$Q_n = \frac{1}{X} \sum_{\ell=1}^{N} x_\ell x_{\ell+n}, \tag{4}$$

where $X$ is defined by,

$$X = \sum_{n=1}^{N} x_n. \tag{5}$$

In Eq. (4) any subindex $n$ of $x_n$ should be understood as $n \pmod{N}$, so that, for example;

$$\begin{aligned}
Q_1 &= \frac{1}{X}(x_1 x_2 + x_2 x_3 + \ldots + x_{N-1} x_N + x_N x_1), \\
Q_2 &= \frac{1}{X}(x_1 x_3 + x_2 x_4 + \ldots + x_{N-1} x_1 + x_N x_2), \\
&\vdots \\
Q_N &= \frac{1}{X}(x_1^2 + x_2^2 + \ldots + x_{N-1}^2 + x_N^2).
\end{aligned} \tag{6}$$

Eq. (4) is obtained from the following consideration: When a sequence of $n$-syllables matches the line definition rule, two conditions have to be satisfied: namely, (a) The starting point has to be the beginning of the word (The probability of this happening after the $\ell$-th syllable in a line is $x_\ell/X$), and (b) The end of the sequence has to be the end of a word, the probability of which is $x_{\ell+n}$. Thus by taking the product of the probabilities of (a) and (b) and summing over $\ell$ we find Eq. (4).

We note that Eq. (4) can be expressed compactly by using the generating function of $x_n$,

$$f_x(z) \equiv \sum_{n=0}^{N-1} x_n z^n, \tag{7}$$

and the generating function of $Q_n$,

$$f_Q(z) \equiv \sum_{n=-(N-1)}^{N-1} Q_n z^n, \tag{8}$$

as follows

$$f_Q(z) \equiv f_x(z) f_x\left(\frac{1}{z}\right). \tag{9}$$

The $Q_n$ distribution given by Eq. (4) has the following properties:

1. Periodicity: $Q_n = Q_{n+N}$.

2. Symmetry within each period: $Q_n = Q_{N-n}$. (This is evident from an identity $f_Q(z) = f_Q(1/z)$ obtained from Eq. (8).)

3. The average value of $Q_n$ is given by the following:

$$\langle Q \rangle \equiv \frac{1}{N} \sum_{n=1}^{N} Q_n = \frac{X}{N} = \langle x \rangle. \tag{10}$$

This is actually a trivial relationship, for $\langle Q \rangle$ is an overall-average probability of a word boundary occuring at a given place.

4. The $Q_N$ distribution has *the highest peaks* at $n = N \times$ (integer), as long as the text has only one type of core metrical line of length $N$. This can be proven as follows: First we note the following inequality:

$$0 \leq \frac{1}{2} \sum_{\ell=1}^{N} (x_\ell - x_{\ell+n})^2 = \sum_{\ell=1}^{N} x_\ell^2 - \sum_{\ell=1}^{N} x_\ell x_{\ell+n}. \tag{11}$$

This implies that

$$Q_N \geq Q_n, \tag{12}$$

which means that $Q_N$ has the largest value. The equality in the above holds if and only if all $x_\ell$ are equal to $x_{\ell+n}$. Therefore, if the $Q_n$ distribution has peaks at $n = k \times$ (integer) of the same height as the peaks at $n = N \times$ (integer), it means that $x_{N-k} = 1$, since $x_N = 1$ by definition. In such a case, we find that the text is actually made of two core metrical lines of length $k$ and length $N - k$. Therefore, given that the text has only one type of core metrical line of length $N$, the $Q_n$ distribution has peaks at $n = N \times$ (integer), which are the highest among all the peaks.

5. (Corollary of 4) Unless the entire text consists of monosyllabic words only. The $Q_N$ distribution has *peaks* at $n = N \times$ (integer). This can be deduced from by extending the above discussion, or from the identity, which can be proven by a straightforward calculation:

$$Q_N - \langle Q \rangle = \frac{1}{X} \sum_{n=1}^{N} (x_n - \langle x \rangle)^2. \tag{13}$$

If $Q_N$ is not a peak, from the property 4 we find that $Q_N = \langle Q \rangle$, which is possible only when all $x_n = \langle x \rangle$ from Eq. (13). Since $x_N = 1$, we conclude then that all words are mono-syllabic.

In the following, we will first calculate the $Q_n$ distribution for two simple model $x_n$ distributions to show that the peak structure and the background is indeed reproduced by Eq. (4). Secondly, we will use Eq. (4) for the actual observed $x_n$ distributions to show that the resulting $Q_n$ distributions match that of the observed distribution.

*Model calculation*

We will begin by calculating the induced $Q_n$ distribution for two cases, the second being a generalization of the first. Both reflect the general features of the actual distribution found.

1. Flat $x_n$ distribution:

$$x_n = \begin{cases} \alpha & \text{if } n = 1 \ldots N-1 \\ 1 & \text{if } n = N. \end{cases} \tag{14}$$

This distribution is plotted in Fig. 21.

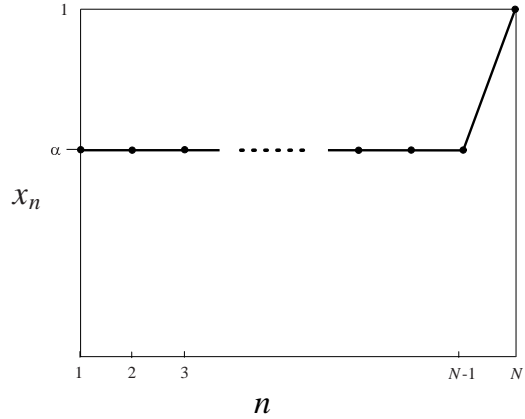

Figure 21: A model distribution of spaces within lines, expressed by Eq. (11).

This distribution models a very simple metrical organisation: if one takes a perfectly randomly segmented text with probability of termination at $\alpha$ and introduces line boundaries at every $N$-th syllable (if necessary), this $Q_n$ distribution is obtained. The average value $\langle Q \rangle$ of the $Q_n$ distribution is given by the following,

$$\langle Q \rangle = \frac{X}{N} = \alpha + \frac{1-\alpha}{N}. \tag{15}$$

The first term of this result comes from the simple termination probability at any position, while the second term comes from the case when the $N$-th position is *not* a word boundary (its probability being $1-\alpha$).

In this case, Eq. (4) yields the following;

$$Q_n = \begin{cases} \frac{1}{X}\left(2\alpha + (N-2)\alpha^2\right) & \text{if } n = 1,\ldots,N-1; \\ \frac{1}{X}\left(1 + (N-1)\alpha^2\right) & \text{if } n = N. \end{cases} \tag{16}$$

The difference between $Q_N$ and the other $Q_n$s is given by:

$$Q_N - Q_i = \frac{1}{X}(1-\alpha)^2 \quad (i = 1, 2, \ldots N-1). \tag{17}$$

This implies that if and only if $\alpha = 1$ we have a flat $Q_n$ distribution, in agreement with properties 4 and 5 discussed above. On the other hand, once $\alpha \neq 1$ the only deviations of the $Q_n$ distribution from the flat distribution are peaks at $n = N \times (\text{Integer})$. Since $X$ is essentially proportional to $N$, the peaks are higher for smaller $N$. In general $N$ is of the order of 10 for most English verse, therefore the height of the peaks tend to be small. For example, if we choose $\alpha = 0.75$ and $N = 10$, we find that $Q_N - Q_i \cong 0.008$, in rough agreement with the observed peak heights in Figs. 7-12.

2. Since in most cases the actual $x_n$ distributions have oscillations with a period value of, we can generalize the above model to introduce an oscillating pattern as follows,

$$x_n = \begin{cases} \alpha & \text{if } n = 1, 3, 5 \ldots N-1; \\ \beta & \text{if } n = 2, 4, 6, \ldots N-2; \\ 1 & \text{if } n = N, \end{cases} \quad (18)$$

where we assume that $N = $ even since it is true for the core isometrical set. Since the $x_n$ distributions in Figs. 14-19 have enhancements at odd $n$'s, we may choose $\alpha > \beta$ in order to model them. The *Hiawatha* data in Fig. 20 may be modelled by choosing $\alpha < \beta$, although other features are more prominent.

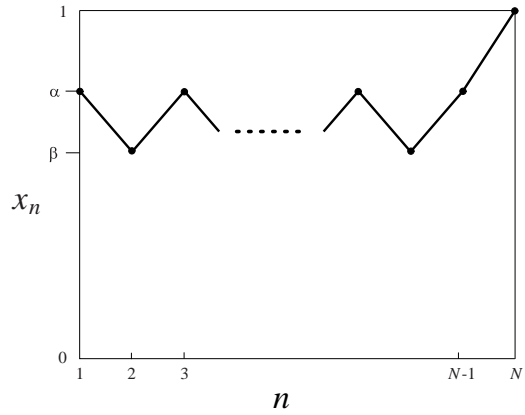

Fig.22: A model distribution of spaces within lines, expressed in Eq. (15).

The average value of the $Q_n$ distribution is given by the following,

$$\langle Q \rangle = \frac{X}{N} = \frac{\alpha + \beta}{2} + \frac{1 - \beta}{N}. \quad (19)$$

This again can be understood very simply as the result of the introduction of a word boundary at the line ends when necessary (with $1 - \beta$ being the probability of the $N$-th position not being the word boundary). In this case Eq. (4) yields the following:

$$Q_n = \begin{cases} \dfrac{1}{X}\left(2\alpha + (N-2)\alpha\beta\right) & \text{if } n = 1,3,5,\ldots N-1; \\ \dfrac{1}{X}\left(2\beta + \dfrac{N}{2}\alpha^2 + \dfrac{N-4}{2}\beta^2\right) & \text{if } n = 2,4,6,\ldots N-2; \\ \dfrac{1}{X}\left(1 + \dfrac{N}{2}\alpha^2 + \dfrac{N-2}{2}\beta^2\right) & \text{if } n = N. \end{cases} \qquad (20)$$

From this we find that the oscillating pattern in the $x_n$ distribution induces oscillating behaviour in the $Q_n$ distribution. In fact, Eq. (22) leads to the following:

$$Q_{\text{even}} - Q_{\text{odd}} = \dfrac{(\alpha - \beta)}{X}\left(\dfrac{N}{2}(\alpha - \beta) - 2(1 - \beta)\right). \qquad (21)$$

In Fig. 23 we give the contour plot of this quantity. As can be seen in this plot, except for the narrow gray area in the $\alpha\beta$-plane, $Q_{\text{even}}$ is greater than $Q_{\text{odd}}$, regardless of whether $\alpha$ is larger or smaller than $\beta$. Furthermore, we readily see that the typical values, say, $\alpha \cong 0.8$, $\beta \cong 0.6$, are very close to the boundary of the gray area, and therefore $Q_{\text{even}} - Q_{\text{odd}}$ is actually very small.

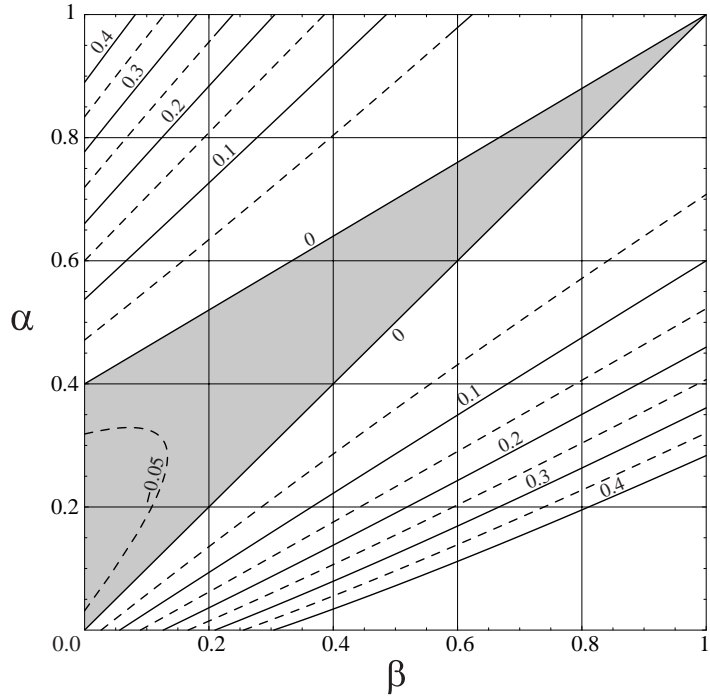

Figure 23: Contour Plot of $Q_{\text{even}} - Q_{\text{odd}}$ given by Eq. (21) for $N = 10$. The values written on each solid line denotes the value of $Q_{\text{even}} - Q_{\text{odd}}$ on that line. The dashed lines represent 0.05, 0.15, and so forth. The shaded region is where $Q_{\text{even}} - Q_{\text{odd}} < 0$.

Furthermore, the $Q_N$ peak height satisfies the following;

$$Q_N - Q_{\text{even}} = \frac{(1-\beta)^2}{X}, \tag{22}$$

This result implies that the $Q_N$ peak is heigher than $Q_{\text{even}}$ unless $\beta = 1$, in agreement with the property 4 discussed above. The result (22) is plotted in Fig. 24 in the same manner as in Fig. 23. Note that typical values, $\alpha \cong 0.8$, $\beta \cong 0.6$, do not necessarily lead to small $Q_N - Q_{\text{even}}$ due to the lack of a gray zone of the type found in Fig. 23. Therefore we conclude that this model explains the general features of prominent peaks at the core metrical line length with possibly much smaller minor peaks at even $n$, as seen in the actual $Q_n$ distributions for Fairfax, Milton, Dryden and Wordsworth.

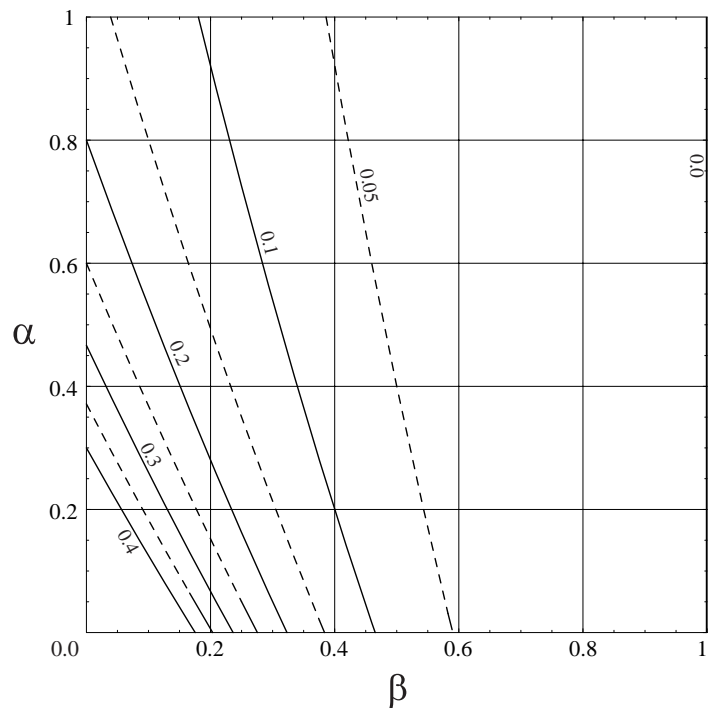

Figure 24: Contour Plot of $Q_N - Q_{\text{even}}$ given by Eq. (22). The values written on each lines denotes the value of $Q_N - Q_{\text{even}}$ on that line. The value $Q_N - Q_{\text{even}} = 1$ is achieved only for $\beta = 1$.

*The reproduction of the observed $Q_n$ distribution*

Eq. (3) can be used to calculate the theoretical $Q_n$ distribution induced by the $x_n$ distribution. Figs. 25 through to 29 plot the actual $Q_n$ distribution (solid line) against the induced distribution for the core isometrical line length (dotted line) for five of the texts listed in Table 1. In each of these cases the plot shows a remarkable coincidence between the actual $Q_n$ distribution and the induced distribution. Furthermore, we note that the peaks, at $n=10$ in Figs. 25 through 28, and at $n=8$ in Fig. 29, are perfectly reproduced by the theoretical distribution. In the case of *Hiawatha* the agreement is still more striking, the theoretical distribution almost perfectly reproducing both the major and minor peaks.

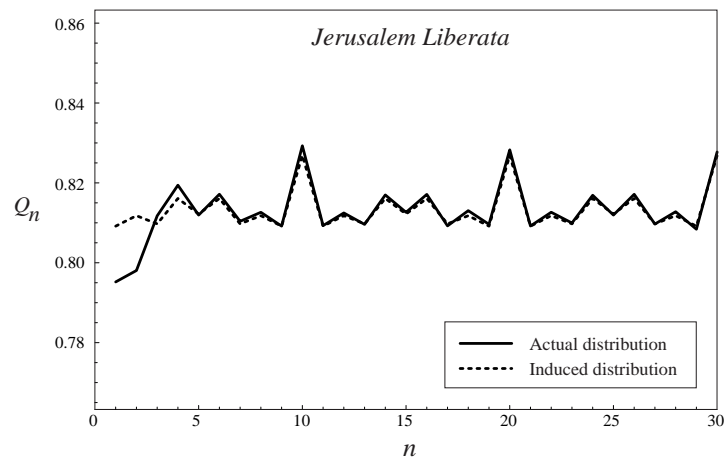

Figure 25: Edward Fairfax, *Jerusalem Liberata*: The actual and the $x_n$-induced $Q_n$ Distributions

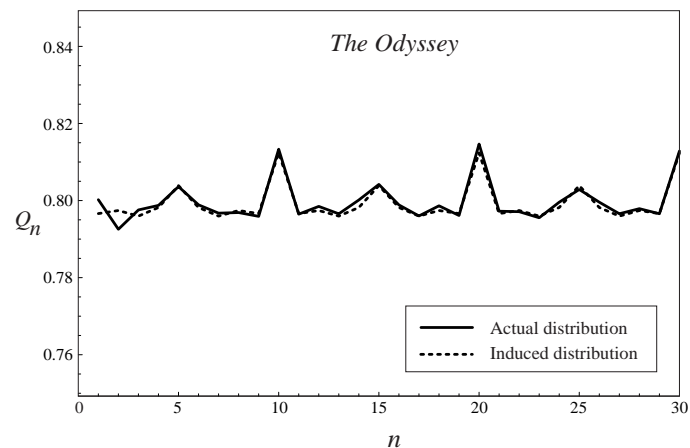

Figure 26: George Chapman, *The Odyssey*: The actual and the $x_n$-induced $Q_n$ Distributions

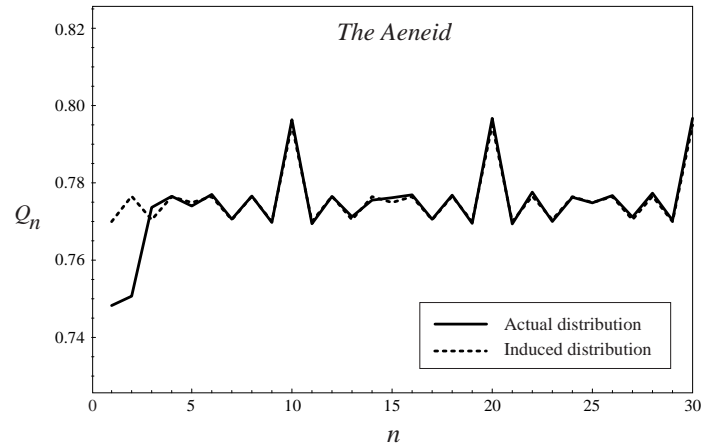

Figure 27: John Dryden, *Aeneid*: The actual and the $x_n$-induced $Q_n$ Distributions

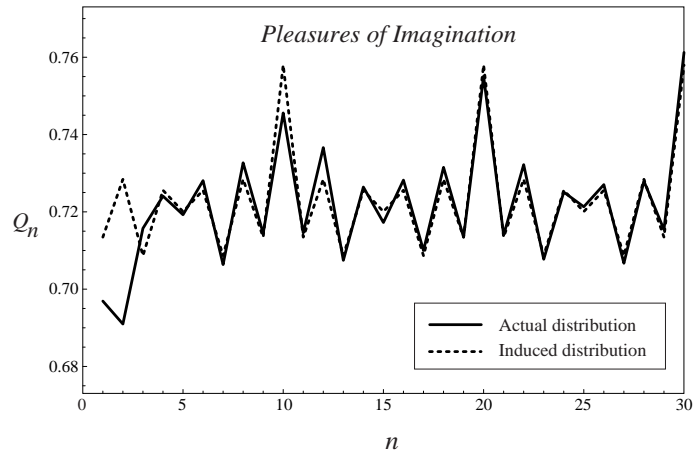

Figure 28: Mark Akenside, *Pleasures of Imagination*: The actual and the $x_n$-induced $Q_n$ Distributions

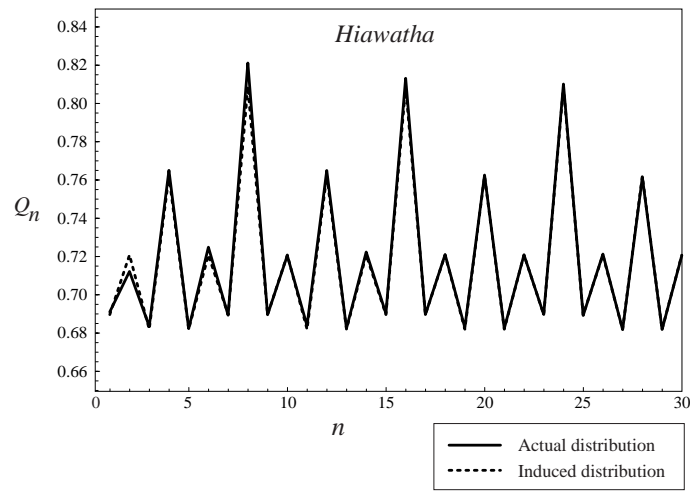

Figure 29: H. W. Longfellow, *The Song of Hiawatha*: The actual and the $x_n$-induced $Q_n$ Distributions

### 3.1.4 Non-Random Sequencing: Syntactic Disruption

We have now shown that there is no evidence of a tendency to use shorter words towards the line end, but that there is a significant tendency to order word length sequencing within lines and that this results from beat patterning. This latter ordering leads to significant subsidiary peaks in the $Q_n$ distribution in addition to the major peaks which are our principal focus of attention. We will now return to these major peaks and consider their significance. Lineation, as has been demonstrated, requires a movement of polysyllables away from the line boundary positions. In other words, the $Q_n$ peak results from a deviation, a non-randomness, in the normal word length sequencing at the line juncture, and any effects within the line are more or less invisible. This is still a real deviation, and we will now turn to consider its significance.

Syntax, properly speaking, is the sequential distribution of language features, and quite understandably, linguists have concentrated their attention on those features, such as the sequence of semantic category items, which are arrayed in an ordered fashion. That is to say it has focused on those features with linguistic significance. Nevertheless, the sequential distribution of word length is as real a feature of syntax as any other, though since it is arrayed randomly it is, for most purposes, of little interest. When, however, we come across an exception, as is the case with isometric verse, the word-length sequencing of which is non-random, a number of rewarding questions arise. Principally, we are led to wonder to what degree the introduction of order into a normally random aspect of syntax affects other, normally ordered, aspects.

To address this issue we need firstly to decide whether there is any significant relationship between word length and part of speech. Since prose output exhibits no apparent order in the word length sequencing at the global level it is reasonable to assume that there is no strong link. However, as noted above, there is some deviation from the random sequencing in the fine structure, namely the depression at $n = 2$ in the $Q_n$ frequency charts. From this we can see that a monosyllable is slightly more likely to be followed by a polysyllable than would be predicted from chance, and grammatical syntax seems the likeliest explanation. That is to say that a word class whose mean length is low is more likely than would be expected from chance to be followed by a class whose length is high. Bearing this in mind it seems probable that detailed examination of the word length characteristics of parts of speech will reveal correlations of interest. However, the effect appears to be weak, and for the time being we will assume that unmetred output presents us with two sequences that are not strongly related. On the one hand a random sequence of word length values, and on the other a highly non-random sequence of parts of speech. Now, if the text is metred and order is introduced into the word length sequence it is clear that there is a *probable* effect on the sequencing of parts of speech, namely that some degree of randomness will be introduced.

This theoretical consideration is open to empirical investigation, and we will offer a hypothesis in this regard. *If two matched English texts are taken, one in prose, one in verse, the sequential distribution of word class items in verse will be closer to a random sequence than that found in prose.* In other words, given a word at position $n$ in a text the frequency totals of word class items in verse will be a better predictor of the class value of the subsequent word at $n+1$ than will be the case in prose (obviously, the frequency totals will be a poor predictor in both cases; we merely suggest that it will be significantly less inaccurate in the case of verse).

In other words we are suggesting that isometric lineation causes disruption in the sequencing of parts of speech, and, moreover, that this disruption does not merely result in alternatively ordered, and convenient, structures, but must constitute a move towards random sequencing. It should be noted that this disruption does not necessarily entail grammatical rule violation or inversion. Neither are necessitated by lineation, though the frequency with which authors resort to them may be increased. The effect to which we are drawing attention is much subtler, and less likely to be salient. Indeed, in carefully written verse it may not be salient at all.

## 3.2 Theoretical Line Violation and Distortion of the Geometric Distribution

We have seen that one way of preventing line boundary transgressions in our artificially lineated prose sample is to alter the word order in non-random ways, and that this would probably result in syntactic distortion. Now suppose that in fact we are concerned to do as little damage to the sense as possible. Assuming that we are already taking as much care as we can in rearranging the syntax, is there anything else that we could do? The answer is yes. We can delete the boundary transgressing polysyllables and replace them with shorter words. That is to say we can introduce more spaces into our text. We can see immediately that this will result in a distortion of the geometric frequency distribution, and that there would be, practically speaking, an increase in monosyllables and a decline in polysyllables in verse.

Empirically, this appears to be case. Constable (1997) has shown that when matched samples of prose and verse are compared the verse has a lower mean word length, and a lower proportion of polysyllables at every length. Fig. 30 charts word length distribution data given in Constable (Forthcoming) for George Eliot's *Middlemarch* (dotted line) and her verse narrative *The Spanish Gypsy* (solid line). The mean word lengths of these two texts are 1.44 and 1.31 syllables per word, respectively.

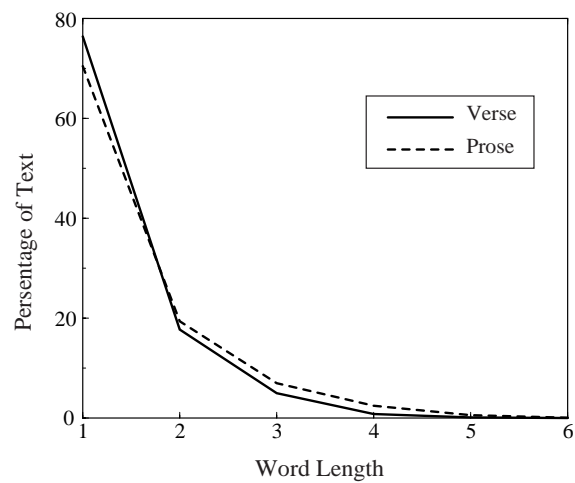

Figure 30: Word Length in George Eliot's *Middlemarch* (prose) and *The Spanish Gypsy* (Verse) Compared

Given the restrictiveness of lineation this tendency to use shorter words is hardly surprising, since it would be very difficult indeed to maintain an approximation to normal syntax if the mean word length were not reduced. Indeed, the ability to reach a satisfactory compromise with the requirements of lineation by passing neither too much nor too little of the burden of restriction over from syntax and onto diction may be a hallmark of the competent verse writer.

These observations may be brought to bear on the report made earlier by Constable (1997: 194), that authors whose mean word length is low in prose tend to reach a much higher proportion of that prose figure in their verse than authors whose mean length in prose is high. It was speculated that the reason for this was that the latter writers were simply more careful of maintaining an approximation to natural syntax. However, it now seems more probable that the reason for this is that writers with low mean word length have fewer polysyllabic word placement problems and are consequently under less pressure to reduce their word length.

## 3.2.1 Variation from the Core Isometrical Line Length

As noted above, the impact of line restriction can be further ameliorated by the use of lines varying from the core isometric length, and in fact very few poems, even in the eighteenth century, are completely isometric with regard to syllable count. That is to say that lines are rarely defined as a rigidly fixed number of syllables in English, lineation being the outcome of a regulation of the distribution of stressed and unstressed syllables to realize beats and offbeats. As noted above, since offbeat positions can occasionally be filled in different ways a number of different line lengths result. For example, in duple verse each offbeat is usually a single syllable, but can on occasion be two syllables, or even left blank. In practice it is rare to find more than two such variants in a line, thus in a work with a core length of ten syllable lines we find a normal maximum of 12 syllables per line, and a minimum of 8, though it is in fact rather rare to find lines of 9 syllables. Neither the motivation underlying the use of variant lines nor their empirical frequencies have been hitherto well understood. In the following discussion we will suggest that the primary motivation is not, as one might think intuitively, the avoidance of sonic monotony, but the amelioration of line length restrictions on the placement of polysyllables. We will begin by discussing the line length distributions themselves, and then examine these frequencies in relation to our data and theory. Fig. 31. gives distribution plots for three poems, Dryden's translation of *The Aeneid*, Milton's *Paradise Lost*, and Wordsworth's *Prelude*, all of which have a core metrical set of ten syllable lines.

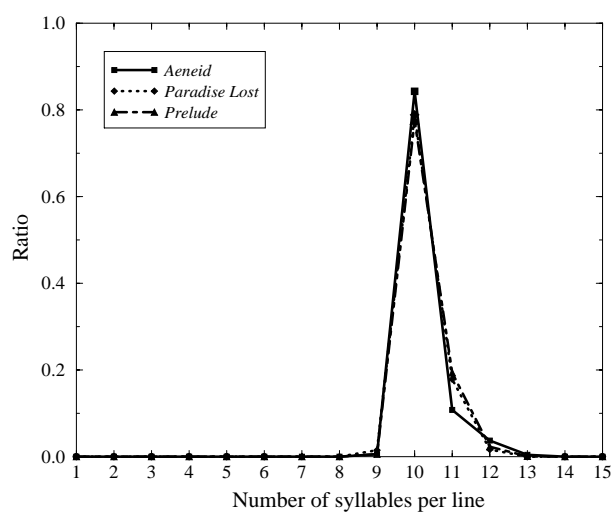

Figure 31: Line Length Frequencies for John Dryden, *The Aeneid,* John Milton, *Paradise Lost*, and William Wordsworth*, The Prelude.*

In these distributions it is evident that when an author employs lines which vary from the core metrical set they tend to employ longer lines rather than shorter lines. This is entirely consistent with the suggestion that variant lines are employed as a result of an attempt to minimize the disruption of lineation, and that when an author varies from the core set it is usually to employ longer words, consequently variant lines are usually longer lines. Further strengthening can be given to this approach by considering the mean word length of the various line length groups, where it has been found longer lines tend to have higher mean word lengths (Constable 1997: 186-187; see also Figs. 14 to 17, and 19 to 20, where the charts show that variant lines usually have a much lower probability of a space at any position). However, some doubt may still remain as to whether the word length characteristics of the variant lines are the simple outcome of internal rhythmic considerations, for example that in order to create an elegant double offbeat authors are more likely to employ a polysyllable. However, consideration of the frequencies of variant lines will suggest that the polysyllabic placement hypothesis is the only plausible hypothesis.

Let us return briefly to the discussion above of the number of theoretical line boundary violations appearing in an arbitrarily segmented text, this time altering our model to incorporate the concept of variant lines. That is to say, suppose that we take George Eliot's *Middlemarch* and arbitrarily segment it into ten syllable lines, *varying up from that length whenever we find our line boundary violating a polysyllable.* Unsurprisingly, the result is a line length profile which coincides almost exactly with the geometric distribution of word lengths, as can be seen if the word length distribution is overlayed on the chart (see Fig. 32). The figures are normalized to facilitate comparison.

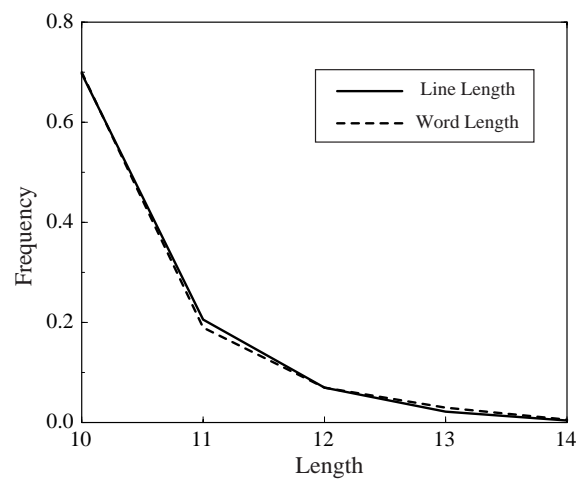

Figure 32: Line Length Frequencies in Artificially Lineated Prose, Compared with Word Length Frequencies: George Eliot's *Middlemarch*.

What we see in such a chart is that the probability of a variant line occurring is simply the probability of a polysyllable violating the line boundary position. Now, if we make the same chart for a representative verse text, comparing it with the word length frequencies of a matched prose sample although we see small and very important differences between the two distributions, the correspondence is still remarkably close. Fig. 33 overlays the line length frequencies (solid line) of Milton's *Paradise Lost* (1667, but begun in the late 1650s) on the word length frequencies (dotted line) of his *History of Britain* (1670, but probably begun in the 1640s and completed in 1655):

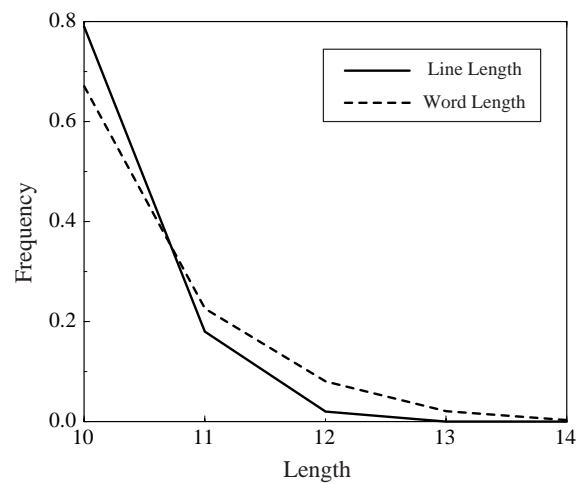

Figure 33: Line Length Frequencies in Verse, John Milton's *Paradise Lost,* Compared with Word Length Frequencies in a Matched Prose Text, *History of Britain.*

In other words, the chart demonstrates that *Milton resorts to variant lines with a frequency that we would expect from the number of theoretical line violations arising in his unmetred language*. The only plausible explanation for this coincidence is that his use of variant lines in verse is motivated by polysyllabic word placement

In the light of these considerations we can see that the more variant lines are employed the less the fact of lineation impinges on the author, and it is in theory possible to use the ratio of actual variants to theoretical violations, discussed above, to judge the degree of flexibility which an author has allowed themselves. For example, in Milton's poem variants are 0.64 of theoretical violations, a figure that we suspect is high. In any case, it seems certain that the variants employed by a composer will hardly ever exceed the number of theoretical violations calculated from a matched prose text (the reader is reminded that these considerations apply to duple verse, and that triple verse may exhibit a different character). As a very rough generalization, we can conclude that since polysyllables are approximately 0.3 of most texts, variant lines are unlikely to approach closely to, and will hardly ever exceed, 0.3 of all lines. It can be further noted that when authors write with a relatively low mean word length, say in the region of 1.2 syllables, per word, they will need to employ variant line lengths much less frequently than authors writing with a relatively high mean word length, 1.3 or 1.4 syllables per word; there are simply less polysyllabic placement problems confronting such an author.

### 3.2.2 Long Lines and Short Lines

The lines specified by metrical rules come in various lengths, the most common lengths being six syllables, in the short lines of ballad stanzas for example, eight syllables, and ten syllables, and these line lengths are expected to have distinctive $Q_n$ distributions with distinctive levels of disruption. Namely, texts lineated using shorter lines will tend to have larger $Q_n$ peaks than text lineated using longer lines. The explanation for this is straightforward, and implied by the mathematical account given above, but some brief comments will make its relevance to compositional questions apparent.

It will readily be seen that if lineation requires the composer to vary from the random word-length sequencing of normal output, then a line system which uses shorter lines, with consequently larger $Q_n$ peaks, will tend to have larger consequences on the sequencing of the text. In other words short lines are more disruptive. Generally speaking, this corresponds to the widespread belief that short lines are less suitable for ambitious content, probably because they are less flexible (Fussell 1965: 450; Tarlinskaia 1993: 137-8). Our account of the $Q_n$ distribution for long and short lines above, allows us to give a clear quantitative account of this intuition.

If we look back, for a moment, to the calculated theoretical violation, it is a simple matter to see that the number of violations is dependent on the line lengths, and that if we examine the relationship between the number of violations, 13,699 for ten syllable lines and 17,123 for eight syllable lines we see that it is given by the relations between the line lengths themselves: 17,123:13,699 = 1.25:1, and 10:8 = 1.25:1. Thus we can see that, at least in regard to this measure of the difficulty caused by lineation, an eight syllable line causes 1.25 times as much work for the writer as a ten syllable line.

## *4. Discussion: Lineation, Poetic Effect, and the Character of Metrical Rules*

We have now described the mathematical character of lineated text, and given a number of empirical illustrations of the facts of the matter. We have shown that the consequences of lineation for a composer are, firstly and fundamentally, syntactical, but that in order to ameliorate this effect an author is likely to use shorter words than they would otherwise have used (this, it should be noted, is an alternative formulation of an argument given in Constable 1997, and Forthcoming b). We have also confirmed that this is true in spite of the use of lines varying from the core isometrical length. Thus we have seen that lineation entails syntactic and dictional disruption.

In conclusion, we would like to attempt to prevent a possible misuse, as we see it, of our position. Since the interaction of syntax and lineation has attracted much attention in criticism (Davie 1955; Baker 1967; Leech 1969: 123-128; Hollander 1975; Ricks 1984; Bradford 1988; Bradford 1993; Wesling 1996) it should be recognized that our approach does not address the issue as traditionally framed, a framing which may be represented by Bradford's lucid presentation in his discussion of Milton:

> The reader must decide whether the typographical format of the verse, the 'white space' at the end of each line, is capable of compelling or modifying syntax or whether it is merely a convention of the printed page. (1988: 187)

Superficially, our work may seem to assist in answering such a question; we have shown that in one sense lineation does indeed "compell" or "modify" syntax. However, while our focus has been on the making of text, critics generally concentrate, as Bradford does, on the reading of it. In other words, they are less interested in whether syntax has been distorted during composition than in whether the introduction of line breaks, as visual phenomena, causes, or *should* cause, the reader to experience the syntax in a different way than they would if there were no lineation (Hartman 1980; Johnson 1990). When put in this form the question is as relevant to free verse as it is to isometrical verse, and therefore lies beyond the scope of our discussion here.

The implications of our findings are numerous, and we would like to touch on two areas of particular interest, firstly the question of poetic effect, and secondly the general character of metrical rules. It should be emphasized that our purpose in these remarks is only to suggest that our findings may assist in opening these areas for further investigation. Both fields are large and highly problematic, and we cannot hope to do more than sketch connections and possible lines of research.

## 4.1 Lineation and Poetic Effect

The possible importance of disruption of diction for our understanding of poetic effect has been outlined in Constable (1997: 197-198), and described in more detail in Constable (Forthcoming b), where the argument is put within the framework of theory of relevance. In the light of our findings above we can recharacterize that disruption, taking into account syntactic disruption, and strengthen the earlier argument. Briefly, it has been suggested that verse form can be regarded as an algorithmic technique for increasing the probability of producing text which simultaneously exhibits intact grammar and ahierarchical implications, and that this can explain the fact that verse form is strongly associated with very rich poetic effects but is neither necessary nor sufficient for their appearance (Buchler 1974: 73).

Communication, it is now widely believed, is not principally a matter of decoding an utterance, but results from a two-stage process of decoding followed by inference construction, the inferences being drawn in accordance with our assumption that a speaker or writer will not require more processing effort of a reader or listener than is merited by the communication (Sperber and Wilson 1995). A crucial part of this process occurs when an individual selects certain of the implications of an utterance, of which there are very large, even infinite, numbers, and decides that these were manifestly intended by the composer to be retrieved by a receiver (to distinguish them from other implications these are referred to as implicatures). The process by which the composer manipulates these retrievals are what we know as style. That is to say, syntactical and dictional choices will tend to structure the implications in certain ways, thus leading the receiver to draw certain conclusions as to which are to be assigned to the category of implicature. Slightly different syntactical and dictional choices will lead to very different conclusions on the part of the reader. Sometimes, often in fact, composers will deliberately arrange for a reader's uncertainty about the strength of an implicature to produce delicate and flexible communicative effects.

The bearing of this on the description of verse given above is straightforward. Verse form forces choice on a number of axes and causes the hierarchy of implications to be to some degree ordered randomly with regard to communicative intent. It will thus, sometimes, be peculiarly difficult to decide which implications are to be assigned to the category of implicature. Such uncertainties do occur in day-to-day circumstances, as for example when we mishear something or misread a word, but, and this is crucial, such cases usually result in extreme incoherence and in grammatical flaws which stimulate the reader either to recover the error or to reject the utterance as irretrievably damaged and undeserving of further interpretative effort. However, the disruptions which occur in high status verse forms are of a slighter, subtler kind, and are not usually accompanied by grammatical damage, though as we have shown lineation alone is sufficient to place it under some strain. Readers are thus unable to construct a satisfactory or coherent implicature, but do not abandon the text, and believing, in accordance with the second principle of relevance (Sperber and Wilson 1995: 260), according to which every "act of ostensive communication communicates a presumption of its own optimal relevance", that the author would not put them to unnecessary labour, will conclude that they have not yet expended sufficient effort to produce a clear interpretation. Consequently, they will dig deeper into the hierarchy of implications in search of a still richer resolution. The process is endless, and with every unsuccessful attempt the reader will, instead of abandoning the project, assume still greater but as yet undiscovered rewards.

Here and there in the literature we find writers who have suggested that the disablements of verse might be fundamental to its effects, rather than peripheral (Ransom [1941] 1996), but these accounts have not been specific as to the character of the restriction, and their linguistics has not permitted a clear explanation of why restrictions should cause readers so often to find verse 'a collection of words that have inexplicable significance, and give one visions and vistas' (Gurney [1916] 1991: 153). We submit that our account of lineation, which is not of course the only restrictive axis in English goes some way to supply this defect. By encouraging the author to sequence grammatical items in a random way, and to use shorter words than they would otherwise have used, lineation weakens the hierarchy of implications, and leads to text which appears to offer a plausible but ever-unfulfilled promise of interpretative rewards.

## 4.2 The General Character of Metrical Rules

The observations in the preceding section put us in a position to offer a remark on the general character of successful metrical rules, a remark which we might have offered earlier but will be seen to be more firmly motivated in this context. As observed by Nowottny (1962: 99-100) there is a considerable difference in reader reaction to patterns determined by "verse form", on the one hand, and, on the other, reactions to "verbal schemes", where by verbal schemes is meant "conspicuous word play [...] repetitions and variations involving the central apparatus of meaning". These latter techniques, Nowottny points out, "are apparently more open to the objection that they impose formalization on the substance of what is said", while verse "however much it may in fact constrain the poet's freedom to put any words he likes anywhere he likes in the poem, easily allows him to preserve the illusion that he has not wantonly sacrificed the meaning to the successful execution of a move in a word-game". Nowottny is here describing the well-known phenomenon of reader resistance to particular kinds of manifest patterning, semantic parallelism for example, a valid point and one that requires explanation. What is particularly puzzling is that readers do not put up equal resistance to verse forms, despite the fact that these forms are as culpably disruptive as verbal schemes. However, in the light of our approach to the effects of lineation and of the character of rich poetic effects a solution offers itself. Verbal schemes are the imposition of further extraneous order on features which are already richly ordered in output, with the result that their consequences are extreme and salient. With lineation the case is, as we have demonstrated, quite different, order being imposed on features which are randomly arrayed, and it's disruptive consequences being subtle and barely noticeable. We think it possible that the principle we have uncovered here will apply elsewhere in English metrical rules, and may have very wide application to the metrical systems of all languages. Namely, we suggest that the metrical rules which become stable and reliably associated with rich poetic effect in any language will tend to regulate a feature which is randomly arrayed in non-metrical output, and thus will tend to introduce, as a byproduct, a degree of randomness into a feature which is usually ordered. By contrast, those which impose further order on previously ordered phenomena will only be regarded as forms of wit, rather than inspiration.